%% file: main.tex
\documentclass[10pt,twocolumn,letterpaper]{article}
\usepackage{iccv}
\usepackage{times}
\usepackage{epsfig}
\usepackage{graphicx}
\usepackage{amsmath}
\usepackage{amssymb}
\usepackage{multirow}
\usepackage{graphicx}
\usepackage[numbers,sort]{natbib} 
\usepackage[table]{xcolor}
\usepackage{tablefootnote}
\usepackage{footnotehyper}
\usepackage[title]{appendix}
\usepackage{pifont}
\usepackage[accsupp]{axessibility} 

\usepackage[pagebackref=true,breaklinks=true,letterpaper=true,colorlinks,bookmarks=false]{hyperref}
\usepackage{cleveref}

\newcommand{\bloss}[1]{{\cal L}_{\rm box}(#1)}
\newcommand{\iouloss}[1]{{\cal L}_{\rm iou}(#1)}

\newcommand{\cmark}{\ding{51}}%
\newcommand{\xmark}{\ding{55}}%

\iccvfinalcopy 
\def\iccvPaperID{3155} 

\ificcvfinal\fi

\begin{document}

\title{Helping Hands: An Object-Aware Ego-Centric Video Recognition Model}

\author{Chuhan Zhang\\
VGG, University of Oxford\\
{\tt\small czhang@robots.ox.ac.uk}
\and
Ankush Gupta\\
Google DeepMind, London\\
{\tt\small ankushgupta@google.com}
\and
Andrew Zisserman\\
VGG, University of Oxford\\
{\tt\small az@robots.ox.ac.uk}
}

\maketitle
\ificcvfinal\thispagestyle{empty}\fi

\urldef{\myurl}\url{https://github.com/Chuhanxx/helping\_hand\_for\_egocentric\_videos}

\begin{abstract}
We introduce an {\em object-aware decoder} for improving the performance of spatio-temporal representations on
ego-centric videos. The key idea is to enhance object-awareness during training by tasking the model to predict
hand positions, object positions, and the semantic label of the objects using paired captions when available. At inference
time the  model only requires RGB frames as inputs, and is able to track and ground objects (although it has not been
trained explicitly for this).

We demonstrate the performance of the object-aware representations learnt by
our model, by: (i) evaluating it for strong transfer, \ie through zero-shot
testing, on a number of downstream video-text retrieval and
classification benchmarks; and (ii) by using the representations learned as input for long-term video understanding tasks (e.g.\ Episodic Memory in Ego4D).
In all cases the performance improves over the state of the art---even compared to networks trained with far larger batch sizes. We also show that by using noisy image-level detection as pseudo-labels in training, the model learns to provide better bounding boxes using video consistency, as well as grounding the words in the associated text descriptions.

Overall, we show that the model can act as a drop-in replacement for an ego-centric video model to improve performance through visual-text grounding\footnote{Code and models available at: \myurl}.
\end{abstract}

\input{sections/intro.tex}   
\input{sections/related_work.tex}   
\input{sections/method.tex}   
\input{sections/implementation.tex}
\input{sections/exps.tex}   
\input{sections/conclusion.tex}   

\clearpage
\vspace{10mm}
\appendix
\noindent{\LARGE \textbf{Appendix}}

\input{supp/box_region}
\input{supp/Implementation}

\input{supp/datasets}

\clearpage

{\small
\bibliographystyle{ieee_fullname}
\bibliography{bib/shortstrings,bib/egbib,bib/refs,bib/vgg_local,bib/vgg_other}
}

\end{document}

%% file: sections/intro.tex
\section{Introduction}
%
In visual-language models there has been a recent move to explicitly build object awareness into the
vision module by adding specialized and bespoke components, or using 
entirely object-centric architectures. The motivation for this partly comes from the attractive
compositional nature of objects and their inter-relationships in language, which enables
inexhaustible novel
combinations~\cite{chomsky2002syntactic,minsky1988society}, and partly from infant cognitive studies
that stress the importance of objects in early visual development~\cite{spelke1992core,lake2017building,tenenbaum2011grow}.
Examples in the video domain include
explicit internal object
representations~\cite{arnab2021unified}, \eg, through
RoI-align~\cite{he2017mask} pooled features either from a pre-trained
region-proposal network
(RPN)~\cite{sun2018actor,saenko2012mid,arnab2021unified,strg}, or from
bounding-box coordinates taken as
input~\cite{orvit,stlt,sthelse,zhang2022object}.
This contrasts with the large body of work where
standard 
representations are learnt end-to-end without any explicit
factorization into objects/entities, such as dual-encoder vision-language models in the image~\cite{clip,jia2021align} and video domains~\cite{bain2021frozen,xu2021videoclip}.

In this paper, we take a different (middle) path and instead use a vanilla video transformer architecture and {\em induce} object-awareness into the video representation by tasking the model to predict object-level properties,
such as their localization and semantic categories, only during training. 

Our target domain is ego-centric video~\cite{epickitchen,grauman2022ego4d},
 and we tailor the object properties used to this. In ego-centric videos 
the actor~\cite{sun2018actor} is often present through their hands, and we therefore task the network to predict both the hands and
the principal objects they interact with. As will be seen, this simple object-aware training boosts the performance
of pre-trained video-language 
architectures significantly, and leads to state-of-art performance across multiple ego-centric benchmarks.
During inference, the model requires only RGB frames as input, and operates as a standard video-language network.

In more detail, our model is built on top of a pre-trained video-language  dual encoder architecture (where there are
separate encoders for the video and text data). We add an additional, but vanilla, transformer decoder head~\cite{vaswani2017attention},
and train with DETR/Mask2former~\cite{detr,cheng2021mask2former} query vectors and  object loss for hands and other objects.  The intuition is that these additional query vectors help the model to attend to and track the hands and salient objects in the scene (these are the `helping hands' of the paper title).
Importantly, we do not require dense frame level ground truth for this training. Rather, we obtain somewhat
\textbf{noisy} and temporally \textbf{sparse} annotations automatically from a hand object detector~\cite{100doh}, and use these to provide prediction targets for the frames where they are available.
This {\em opportunistic} use of annotations is pragmatic as object detectors trained on third-person datasets
(such as COCO) do not perform so well on the ego-centric domain, where the scenes are more crowded and
objects are often small and can be motion blurred. By only requiring annotations for a subset of frames, where they can be reliably produced automatically, \textbf{we are able to train on large-scale data without requiring expensive manual supervision.}

Although we train with noisy and sparse image-level object localization, our model can learn to predict better and denser bounding-box trajectories through large-scale training due to the spatio-temporal consistency which naturally presents in videos.  Also, it is able to predict semantic grounding by learning to map the object appearance to the nouns in the video captions. 

It is worth noting that we are using hand detectors because hands are a common and important object in ego-centric videos. However, the object-centric method we are proposing has greater scope than ego-centric videos and can be applied to other scenarios with other object types providing the `helping-hand'.

In summary, we make the following contributions:
(i)~We propose a method to induce object-awareness in video-language models for an architecture composed of standard neural modules. The model 
only requires RGB frames as inputs, and thus is  a drop-in replacement for any ego-centric video model.

(ii)~The model can be trained opportunistically using available and sparse frame-level and noisy annotations, produced
automatically. 

(iii)~We demonstrate state-of-the-art strong (zero-shot) transfer for cross-modal retrieval 
to other ego-centric datasets namely, EpicKitchens-MIR and EGTEA improving prior art by 2-4\%.

(iv) We evaluate the
grounding quantitatively using the EpicKitchens-VISOR
dataset~\cite{VISOR2022,epickitchen} and find that the model outperforms the base hand-object detector used for training supervision.

(v) Finally, we also demonstrate that the representations learned can be used as input in long-term video understanding tasks like EgoNLQ and EgoMQ. The objectiveness in the representation helps the model outperform other models trained on the same training set on these two tasks.

%% file: sections/related_work.tex
\section{Related Work}

\paragraph{Vision and Language Representation Learning.}
Different from transferring representations learned for classification on a 
fixed set of object categories~\cite{Krizhevsky12,Simonyan15}, recent 
vision-language pre-training (VLP) works leverage large-scale supervision from 
free-from text descriptions of images and videos.
These methods use image captions~\cite{clip} or video 
sub-titles~\cite{miech2019howto100m,xu2021videoclip} with either independent
dual encoders for the visual and text 
modalities~\cite{luo2022clip4clip,jia2021align,bain2021frozen}, or via joint 
encoders with cross-attention across 
modalities~\cite{alayrac2022flamingo,li2022blip,li2021albef,yuan2021florence}.
We also use dual-encoders which are kept frozen due to compute limitations.
To explicitly build-in object-awareness into \emph{image} representations,
object-level features extracted from pre-trained object-detectors are aligned
with the text descriptions~\cite{li2020oscar,zhang2021vinvl,tan2019lxmert,lu2019vilbert,chen2020uniter,zhong2022regionclip}
The object-level text alignment is further augmented with the object-box 
prediction task for grounding in~\cite{zeng2021xvlm,dou2022fiber,zhang2022glipv2,kamath2021mdetr}.
\textit{VLP for ego-centric videos} has recently been explored~\cite{egovlp,zhao2022lavila} 
to bridge the domain gap between representations learned from third-person videos
found commonly~\cite{miech2019howto100m}, and first-person ego-centric videos.
We further extend image based object-aware VLP methods to \emph{ego-centric videos} by training to predict the auto-generated hand-object boxes extracted from pre-trained detectors~\cite{100doh}, while
requiring only RGB input during inference, making our model a drop-in replacement
for ego-centric VLP models albeit with enhanced object-awareness.

\begin{figure*}[t]
    \centering
    \includegraphics[width=0.9\textwidth]{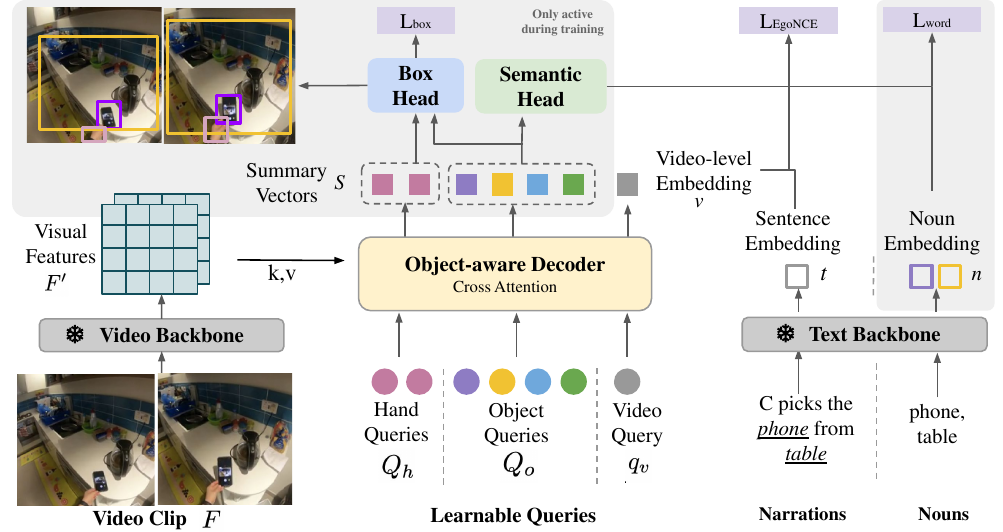}
    \label{fig:arch}
    \vspace{2mm}
    \caption{\textbf{The object-aware vision-language model architecture.} The architecture is made up of three parts: A video backbone, a text backbone and a object-aware decoder. The decoder is a cross-attention transformer, it takes the visual feature map $F'$ as keys and values, which are attended by a set of learnable queries. In these permutation-invariant query vectors, the hand and object queries $Q_h, Q_o$ are trained to be \textit{object-aware} and predict the localization and class of hands and objects. The video query $q_v$ attends to both the visual feature map through cross-attention layers, and the object feature map through self-attention layers in the Transformer decoder and output a video-level embedding $v$ to be matched with the text embedding $t$.}
    \end{figure*}

\paragraph{Weakly Supervised Video Text Grounding.}
A key challenge for grounding in videos is the lack of large-scale object-level annotations for videos.
While such annotations are readily available for  synthetic datasets~\cite{yi2019clevrer}, expensive manual annotation is required for real videos.

Hence, weakly supervised method have been developed which leverage the video sub-titles/descriptions to map nouns/verbs to regions in frames. This is typically achieved by first extracting bounding-box/segmentation regions from pre-trained detectors for objects and humans, and aligning them with keywords using max-margin~\cite{zhou2018weakly} or constrastive~\cite{li2021weakly} objectives. Similarly, \cite{mavroudi2022weakly} also align words in the video captions to regions from pre-trained RPN by modelling the region-word associations as latent variables of a conditional generative variational auto-encoder. More recently, \cite{comma} use cross-attention across text and candidate regions, and find (soft-) associations based on attention weights. While our model similarly absorbs object bounding-boxes and category information obtained from pre-trained object and hand detectors during training to induce object-aware representations, these detectors are not required during inference.

\paragraph{Object-Oriented Learning in Videos.}
Most vision tasks for images involve objects~\cite{lin2014microsoft,krishna2017visual,krishna2018referring,johnson2016densecap}. In videos, there are also a broad range of object-oriented tasks: some works treat learning object-level information as an end task, they design models to predict object bounding-boxes and masks~\cite{xu2018youtube,yang2019video,kristan2021ninth,zhang2022fine,liu2022joint,liu2020forecasting,bambach2015lending}; other works use object knowledge to achieve some other high-level tasks, for instance,  object-centric scene generation~\cite{yang2021objectnerf,kundu2022panoptic} and decomposition~\cite{henderson20neurips,singh2022simple,elsayed2022savipp}, and action recognition~\cite{strg,herzig2019spatio,orvit,avraham2022bringing,zhang2022object,zhou2023can,pirsiavash2012detecting}.
Our method uses object information to assist vision-language alignment in egocentric videos, so that the model learns grounded video representations that can generalize better.
SViT~\cite{avraham2022bringing} uses object queries shared between images and videos in order to predict hand-object bounding boxes in videos, whilst only requiring image-level supervision during training. However, the object queries are not used for vision-language alignment.
Our previous work~\cite{zhang2022object} showed that encoding object-level information in the model helps transfer learning in various video understanding tasks, but the model required GT bounding boxes as input during inference.
In contrast, in this work we model does not require this information during training, and we show box prediction and vision-language alignment can be combined and benefit both in-domain and out-of-domain tasks.

%% file: sections/method.tex
\section{Object-Aware Vision-Language Learning}
We first describe the task of object-aware vision-language learning and the architecture of our model. These are followed by the training objectives for vision-language matching and weakly-supervised text grounding. 

\subsection{Overview}
In vision-language representation learning the training data consists of short video clips (RGB frames) and an associated free-form text string containing words that describe the visual content.
Typically, dual encoders are trained on this paired data, with a visual-encoder that ingests the video clip and a language-encoder that ingests the text~\cite{clip,xu2021videoclip,luo2022clip4clip,egovlp,bain2021frozen}. The dual encoders are trained with a contrastive objective~\cite{infonce} such that the cosine similarity for matching vision-text pairs is optimized to be higher than the similarity of negative/not-matching pairs. This pre-training objective enables evaluation on downstream vision-language tasks like video-text retrieval and action classification in a zero-shot manner~\cite{clip}.

Our object-aware model follows the data, training and evaluation pipeline as above, except that the model is also tasked to output object-level information (e.g., bounding boxes and object categories) {\em during training}. By tasking the model to predict object bounding boxes and names which can be matched to the nouns in the narration, the model learns grounded and fine-grained correspondence between the modalities. The object-level prediction is used as an auxiliary objective in training but not used at inference time. 

In more detail, as shown in figure~\ref{fig:arch}, there are two types of object-level prediction: (a)~hand and object bounding boxes; and (b)~object names. 
Since ground truth of boxes and object names are not available, and most traditional detectors~\cite{detr,zhou2022detic} fail to identify objects well in ego-centric videos, we cannot rely on strong supervision for the predictions. Instead, we generate bounding box targets (for the hands and other objects) using a robust off-the-shelf hand-object detector~\cite{100doh}, though these targets will only be available for some of the frames and are noisy. While for object name prediction, we use a weakly supervised method to align the predicted names with nouns in the paired narration (detailed in section~\ref{sec:class_pred}). In both cases, the supervision is {\em opportunistic} and only applied when available.

\subsection{Architecture}
\paragraph{Dual Encoder.}
We use dual encoders as our visual and text backbone for efficiency. 
The visual encoder ingests a video clip $F$ of RGB frames ${F} = (f_1, f_2, 
\hdots, f_T)$, where $f_i \in \mathbb{R}^{ H{\times}W{\times}3}$ and $T$ is the number of frames.
The clip ${F}$ is encoded by a Video Transformer~\cite{timesformer} which tokenizes the frames by 3D patches to produce a spatially downsampled feature map  ${F'} = (f'_1, f'_2, 
\hdots, f'_T)$, where $f'_i \in \mathbb{R}^{ H'{\times}W'{\times}C}$.  It outputs a visual embedding $v \in \mathbb{R}^{C}$.

The text encoder is a Transformer~\cite{vaswani2017attention} that inputs words tokenized by a BPE tokenizer~\cite{gage1994new}. It encodes two type of inputs: (a)~a narration sentence which describe the contents of a clip; and (b)~a noun set which contains noun phrases from the narration sentence.  At the output, the embedding corresponding to the EOS token is taken to be the embedding for the full sentence $t \in \mathbb{R}^{C}$, and multiple noun embeddings $n \in \mathbb{R}^{C}$.

\paragraph{Object-Aware Module.}
The object-aware module is a cross-attention Transformer which has a permutation-invariant set of learnable vectors as queries (similar to DETR~\cite{detr} and Mask2former~\cite{cheng2021mask2former}). The queries are at video-level, shared between frames. They consist of three sets: two hand queries ${Q_h} = (q_{h1}, q_{h2})$ for the left and right hands; $K$ object queries ${Q_o} = (q_{o1}, q_{o2} \hdots, q_{ok})$; and a video-level query ${q_v}$. These queries are learned, and attend to the visual feature map ${F'}$  from the visual backbone and output a set of summary vectors  ${S} = (s_{h1}, s_{h2}  ;  s_{o1},\hdots s_{ok})$ corresponding to each input query.

The object-aware module operates on the visual content without any interaction with the text information. It consists of six cross-attention blocks. As in a traditional Transformer decoder~\cite{vaswani2017attention}, in each block, there is a multi-head self-attention layer and a  multi-head cross-attention layer. The self-attention layer enables interactions between hand, object and video queries, and the cross-attention layer allows the query to extract object-oriented information from the visual content.
\paragraph{Bounding Box Head.}
The hand query vectors $Q_h$ and object query vectors $Q_o$ are trained to predict the bounding box of the hands and objects respectively in each frame. Note these query vectors and summary vectors are at the video level; to predict boxes at frame level, we condition a summary vector $s_j$  of object j on a learnable frame index vector $x_{i}$ by concatenation of $s_j$ and $x_i$, and use a multi-layer perceptron $F_{box}$ to project them onto a bounding box $\hat{b}_{j,1}$, where $i$ is the frame number:
\begin{equation}
 \hat{b}_{ji} = F_{box}(s_j;x_i)
 \label{eq:concat}
\end{equation}
As a result, we will have a time series of bounding boxes  $(\hat{b}_{j,1}, \hat{b}_{j,2}, \hdots \hat{b}_{j,T})$ from each $j^{\text{th}}$ summary vector.

\paragraph{Semantic Head.}
We assign semantic meanings to object summary vectors, standing for the object name/class. To achieve this, we project $s_j$ onto a word embedding $\hat{n}$ with a multi-layer perception $F_{semantic}$ :
\begin{equation}
 \hat{n} = F_{semantic}(s_j)
\end{equation}

\subsection{Training Objectives}
\paragraph{Vision-Text Matching.}
We follow EgoVLP~\cite{egovlp} and use EgoNCE loss as the objective for matching between video-level embedding $v$ and sentence-level embedding $t$ of the narration. In one batch $\mathcal{B}$, the positive sample set $\mathcal{P}_m$ is made up of a sample $i$ and other samples that share at least one noun and one verb with it: $\mathcal{P}_m=\{n\in \mathcal{B}~|~\text{noun}(n)\cap\text{noun}(m)\neq\varnothing, \text{verb}(n)\cap\text{verb}(m)\neq\varnothing\}$. And for each sample $i$, there is a hard negative sample $i'$ sampled from the same video. Hence, the samples in the original batch $\mathcal{B}$ and 
their hard negative counterparts together form the new batch $\mathcal{\widetilde{B}}$.

The objective matching video-to-text (v2t) for a video embedding $v$ is formulated as below; in practice the symmetric text-to-video (t2v) matching objective is also used (omitted for brevity):
\begin{equation}
	\mathcal{L}^\text{ego}_\text{v2t}=\frac{1}{| \mathcal{P}_m |}\sum_{k\in\mathcal{P}_m}  \log 
	\frac{
	\exp(v^Tt_k /\tau)
	}
	{  \sum_{n\in \mathcal{\hat{B}}} \left( \exp(v^Tt_n /\tau) +
	\exp(v^Tt_{n'} /\tau)\right) 
	}.
	\label{egonce}
\end{equation}

\paragraph{Bounding Box Prediction.}
We use the 100DOH off-the-shelf hand object detector~\cite{100doh} to produce bounding boxes of two classes on each frame as supervision: hand and objects that are in contact with hands. 

There are two challenges in using the detections for training supervision: 1)~the image detector acts at the image level independently and does not provide box-ID association over different frames in a clip; 2)~many hands and objects are missed due to motion blur and domain gap in ego-centric videos.
Therefore, we apply Hungarian matching between predicted boxes $\hat{b}$  and ground-truth boxes $b_{i}$ on single frames independently, so that for each $b_{i}$, we find the matched prediction $\hat{b}_{\sigma(i)}$ to minimize the global matching cost. The final loss on bounding boxes is computed as the sum of the $\ell_1$ loss and the Generalized IoU loss $L_{iou}$ ~\cite{rezatofighi2019generalized} on paired boxes:
\begin{equation}
\bloss{b_{i}, \hat{b}_{\sigma(i)}} = \iouloss{b_{i}, \hat{b}_{\sigma(i)}} + ||b_{i}- \hat{b}_{\sigma(i)}||_1
\end{equation}
and, to tackle the problem of missing objects, we do not penalize boxes that are not matched to nouns unlike traditional detection tasks.

\begin{figure}[ht]
\centering
\includegraphics[width=\linewidth]{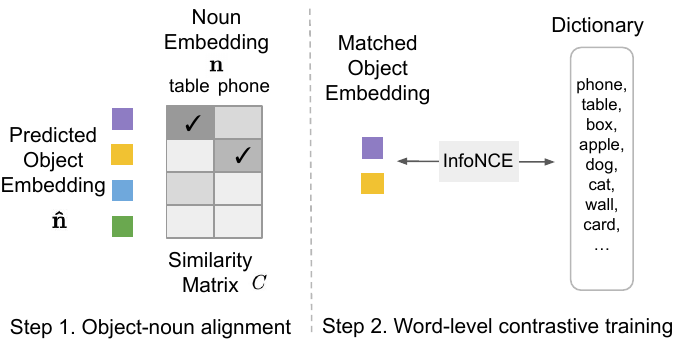}
\caption{\textbf{Training the model to predict object classes.} \textbf{Left: Object-noun alignment.} First, the nouns in the video descriptions are matched against the predicted classes using Hungarian matching, to choose the most matched summary vectors. \textbf{Right: Word-level contrastive training.} Next, we supervise the matched predicted classes using a contrastive objective~\cite{infonce} against all the nouns in Ego4D taxonomy, to have similar embeddings as the corresponding nouns.}
\label{fig:word_loss}
\vspace{-3mm}
\end{figure}

\paragraph{Object Class Prediction.} \label{sec:class_pred}
We have noun embeddings from the video description and a set of predicted object name embeddings from the summary vectors, the task is to find the correspondence between them so that we can use the ground-truth nouns from the description as supervision for the predicted object names.
As shown in figure~\ref{fig:word_loss}, we align the nouns in the narrations and the names of object-boxes in two steps:

\noindent (1) \textbf{Object-noun alignment:} We score the predicted object name embeddings $\hat{n}$ against the ground-truth noun embeddings ${n}$ to construct a similarity matrix $C\in \mathbb{R}^{K\times N}$, where $K$ is the number of object queries and $N$ is the number of noun phrases in the description as following:
\begin{equation}
C({n},  \hat{n}) = \frac{n \cdot  \hat{n}} {||n|| || \hat{n}||}
\label{eq:align_nouns}
\end{equation}
Cost matrix $-C$ is used in Hungarian matching to select the matched summary vector for each noun phrase.

\noindent (2) \textbf{Word-level contrastive training:} We apply InfoNCE loss on the matched embeddings $\hat{n}_j$ and $n_j$  against the embeddings of all the nouns $n'_k$ in Ego4D taxonomy dictionary $\mathcal{D}$~\cite{grauman2022ego4d}:
\begin{equation}
\mathcal{L_{\text{word}}}= -\frac{1}{N}\sum_{j=1}^{N} \log\frac{\exp(\hat{n}_j^T n_j / \tau)}{\sum_{{k \in\mathcal{D}}}\exp(\hat{n}_j^T n'_k / \tau)}
\label{eq:word_loss}
\end{equation}

\paragraph{Training Loss.}
The total training objective is the sum of the vision-text matching loss and the auxiliary losses on object vectors:
\begin{equation}
\mathcal{L_{\text{total}}}= \mathcal{L}^\text{ego}_\text{v2t} +  \mathcal{L}^\text{ego}_\text{t2v} + \mathcal{L}_\text{box} + \lambda_{word}\mathcal{L}_\text{word}
\label{eq:final_loss}
\end{equation}

\subsection{Inference}
Once trained, the model acts as a standard ego-centric vision-language video model which operates just on video frames and text descriptions, without requiring further access to object boxes or detectors.
However, if desired, hand and object box detections and names can be read out for each frame at inference using the summary vectors, which can be used for grounding the input text description.

\begin{figure*}[ht]
\centering
\includegraphics[width=\textwidth]{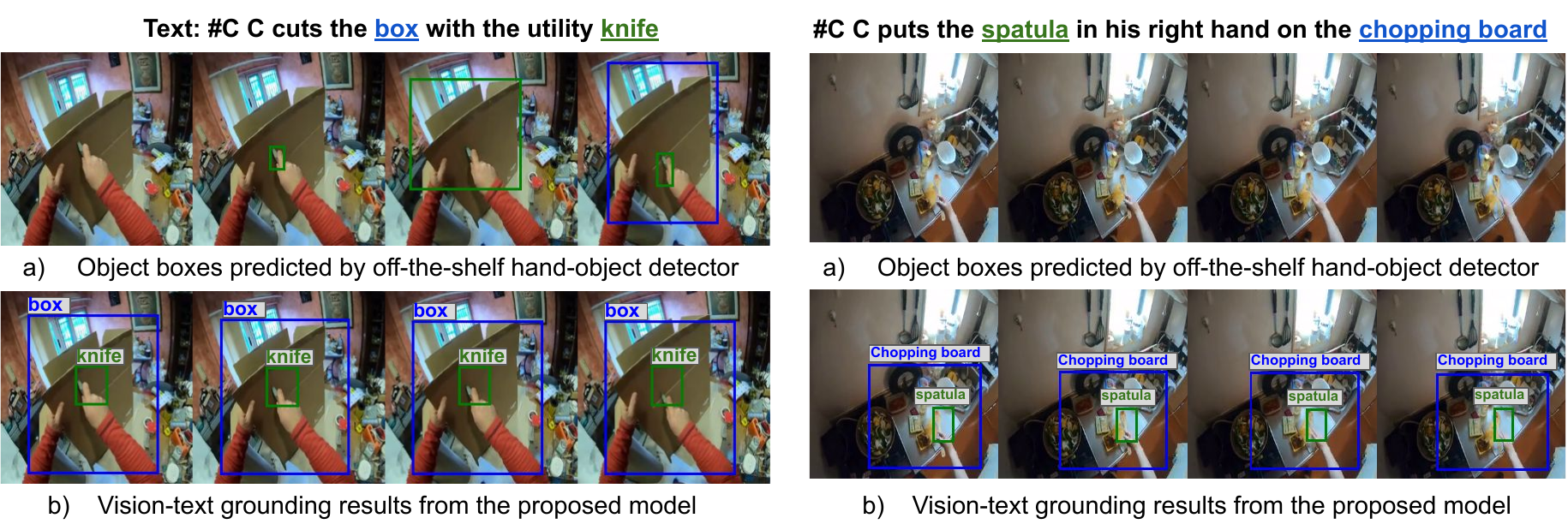}
\caption{\textbf{Visualization of text grounding on EgoClip.}  We show the comparison between detections from off-the-shelf 100DOH hand-object detector~\cite{100doh} (used for training supervision) and the predicted boxes from our model respectively. \textbf{ (a)} The detections are noisy, objects are missed, and there is no temporal association of the detected boxes across frames. \textbf{(b)} The trained model learns temporally consistent tracks as well as object categorization using only noisy frame-level box-supervision and weak supervision from the texts.}
\label{fig:viz}
\vspace{-3mm}
\end{figure*}

%% file: sections/implementation.tex
\section{Implementation}
In~\cref{sec:weak_sup}, details of extracting the hand and object detections from 100DOH pre-trained detector are summarized, followed by the architectural details in~\cref{sec:arch_impl}. Finally, in~\cref{sec:train_details}
the training pipeline, model and input specification, and optimization details are provided. More details can be found in the supplementary materials.

\subsection{Weak Supervision from Pre-trained Detector}\label{sec:weak_sup}
We uniformly sample four frame from each clip in the EgoClip dataset~\cite{egovlp} as the input to the 100DOH hand-object detector~\cite{100doh}. The short side of the frame is resized to 640 pixels. There are 16~million 
frames in total, and the average number of boxes detected per frame is 1.96 for hand and 1.67 for object. Among all the frames, about 15.8\% frames have no hands detected and 17.9\% frames have no object detected. The average size of hand boxes is 2.8\% of the frame size, while the average size of object boxes is 19.4\% of the frame size. We use the top 2 hands and top 4 objects detected in the scene as supervision.

\subsection{Architecture}\label{sec:arch_impl}
The object-aware module is a 6-layer cross-attention transformer with 8 attention-heads in each layer. The hidden dimension in cross-attention layers is 768, and the video embedding, object embeddings and text embeddings are projected to 256 dimensions before computing the cosine similarity score. We set the number of hand queries to 2, number of object queries to 12, which is designed to be larger than the maximum number of objects in the supervision. We use TimeSformer-L (TSF-L)~\cite{timesformer} as the visual encoder, and a 12-layer Transformer~\cite{vaswani2017attention} from LaViLa~\cite{zhao2022lavila} as the text encoder. 

\subsection{Training Details}\label{sec:train_details}
In training, we uniformly sample $4{\times}224{\times}224$ RGB frames from each clip.
Hand and object boxes are pre-extracted from these 4 frames using off-the-shelf detector and used as supervision. We keep the visual and text encoder frozen in training. Only the object-aware decoder, query vectors, and the MLP projection layer on text embedding parameters are learned during training. 
We train the model for 5 epochs on one A6000 GPU, with batch size 128. We use AdamW~\cite{adamw} as the optimizer and set the learning rate to $3e{-}5$. $\lambda_{word}$ is set to 0.5 to balance the scale of the four losses.

%% file: sections/exps.tex
\section{Experiments}
\Cref{sec:exp-datasets} introduces all the datasets we used for training and evaluation,
followed by the evaluation protocols in~\cref{sec:eval-protocol}. Finally, we discuss the ablation studies (\cref{exp:ablation}) and compare to prior SOTA methods on different benchmarks (\cref{exp:sota}). 

\subsection{Datasets}\label{sec:exp-datasets}
\paragraph{Ego4D/EgoNLQ/EgoMQ/EgoClip/EgoMCQ~\cite{grauman2022ego4d,egovlp}.} Ego4D is a massive-scale dataset focusing on ego-centric videos. It contains 3670 hours video for many different tasks, including action anticipation, AV diarization, etc. EgoNLQ and EgoMQ is a subset for natural language queries and moment query, designed for testing the models' episodic memory and long-term video understanding. \cite{egovlp} proposes  a new subset EgoClip for vision-language pre-training, comprising 3.8M clip-text pairs. They also introduce EgoMCQ (\ie, Egocentric Multiple-Choices-Question) as a downstream evaluation dataset for the pre-training. Given a text query, the model tasked to choose the paired video clip from 5 candidates.  The evaluation metrics is `intra-video' and `inter-video' accuracy, depending on where the candidates are chosen from.\\

\noindent\textbf{Epic-Kitchens-MIR~\cite{epickitchen}.} Epic-Kitchens is a large-scale ego-centric dataset with 100-hour activities recorded in kitchens.  Epic-Kitchens-MIR is a subset with about 9881 clip-text pairs for vision-language retrieval. It is designed for multi-instance retrieval. The model is evaluated on retrieving the paired text/video given a query text/video. The evaluation metrics are mean average Precision (mAP) and normalized Discounted Cumulative Gain (nDCG).\\

\noindent\textbf{VISOR~\cite{VISOR2022}} is a benchmark built on Epic-Kitchens for segmenting hands and active objects in egocentric video.  It has pixel-level annotations covering 36 hours of untrimmed video and 257 object classes. We utilize this annotations in its val split for evaluations on vision-text grounding.\\

\noindent\textbf{EGTEA~\cite{li2018egtea}} contains 28 hours of cooking activities from 86 unique sessions of 32 subjects. Fine-grained actions are classified into 106 classes. We retrieve the text-descriptions of action classes to evaluate the model for action classification on the test set of all three splits. We measure the performance using mean-class accuracy  and top1 accuracy.

\subsection{Evaluation Protocol}\label{sec:eval-protocol}
We evaluate the performance of our model in the following three aspects:\\
\noindent\textbf{Zero-shot transfer.} To test the transferability and generalization ability, we conduct zero-shot evaluation on multiple-choice questions (EgoMCQ), multi-instance retrieval (Epic-Kitchens-MIR), action classification (EGTEA). Among these datasets, videos in EgoMCQ are from the same data source as our pre-training dataset Ego4D. Other datasets demonstrate a domain gap, hence, evaluate for transferable representations. 

\vspace{2mm} 
\noindent\textbf{Episodic memory.} To evaluate the richness of the representations learned by our model, we use the video representations to solve Episodic memory tasks in Ego4D. 
Following ~\cite{egovlp,zhao2022lavila}, we pre-extract the video features from our model first. Using these pre-computed features as input, we train a VSLNet~\cite{zhang2021vslnet} for temporal localization in NLQ, and train a VSGN~\cite{zhao2021vsgn} for moment retrieval in MQ.

\noindent\textbf{Vision-language grounding.}
Due to the lack of ground-truth for object-grounding in EgoClip,  we evaluate the grounding ability of the model on VISOR instead. VISOR is an egocentric dataset with scenes in the kitchens, where frames are annotated sparsely with segmentation masks and object names. We re-propose the manually annotated segmentation masks in it to extract ground-truth bounding-boxes for hands and in-contact objects.
To carry out the zero-shot evaluation, we take the annotated frames in the val split, filter out the not-in-contact objects in each frame, and convert all the segmentation masks to bounding boxes as ground truth.
The predicted object boxes are matched with ground-truth object boxes using noun alignment as during training (\cref{eq:align_nouns}), while the left/right hands are predicted from the first and the second hand queries respectively. 
We repeat a single frame 4 times temporally, and resize it to 224 $\times$ 224 pixels to be consistent with the pre-training resolution.
The predicted bounding-boxes are evaluated to be correct if their centers lie inside the ground-truth bounding-boxes.

\subsection{Ablations}\label{exp:ablation}
\paragraph{Losses.} We ablate the combination of losses on three downstream benchmarks in \cref{tab:loss}. Results showing that having both  $L_{box}$ and  $L_{word}$ leads to the best performance. With the same architecture, when training the model using only $L_{Ego}$ without introducing any object-awareness, the performance is ~2\% lower on EK100-MIR and EGTEA compared to the object-aware one. 
\input{tables/loss.tex}

\input{tables/box_ablate}
\input{tables/n_query.tex}

\input{tables/sota.tex}

\paragraph{\textbf{Quality of detected boxes.}}
The extent to which a model can acquire object-level information is constrained by the quality of the bounding boxes from the off-the-shelf detector. To investigate how much the quality of boxes affects our training, we detect hand and object boxes on EgoClip training set using 100DOH~\cite{100doh} with 256p and 640p images as input -- with larger image size, the objects should be more precisely delineated. We use the two sets of boxes as supervision in our training and show results in \cref{tab:better_box}. Even when training with noisy boxes from 256p, our model is better than the previous SOTA model. When boxes from 640p are used, the averaged zero-shot transfer performance is furthur improved by 1\% on Epic-Kitchens and EGTEA, showing that our method can bring larger improvement over the non-object-aware method when given better boxes. 
Furthermore, a significant boost on the grounding results is also observed on VISOR when using boxes with better quality. 

\paragraph{Number of object queries.}
We use different number of query vectors in the object-aware decoder to see its impact on both vision-language tasks and the grounding task. We show zero-shot transfer results in \cref{tab:nquery} and vision-text grounding results in \cref{tab:visor}. The number of queries has relatively small impact on vision-language representation learning, the performance gaps on zero-shot transfer tasks are mostly small than 1\%. However, a smaller number of object queries leads to much better results for in-contact object grounding on VISOR, 4 queries is better than 12 objects by 3.6\% in localization accuracy. The reason is that too many queries result in a large number of predicted boxes, which increases the probability of mis-matching.

\paragraph{\textbf{Impact of hand boxes.}}
We ablate the impact of having hand boxes as supervision in our training. Results are shown in \cref{tab:hand_box}, training the model to predict hand as well as objects help the model to get about 1\% higher  zero-shot transfer performance on EK and EGTEA.
\subsection{Comparison to the SOTA}\label{exp:sota}
\paragraph{Zero-shot transfer.}
\input{tables/hand_ablate}

We compare to previous SOTA in \cref{tab:sota}. Our model is comparable on EgoMCQ and better on EK-MIR and EGTEA, showing its good zero-shot transferability.
Due to limited compute resources, we are not able to unfreeze the visual backbone to train end-to-end or increase the batch size further. Despite these disadvantages, our method outperforms the previous SOTA on two tasks for models that have been trained end-to-end.

The main difference between LaViLa(L) and ours is the object-aware training and hard negative sampling; 
Ours\dag  \hspace{1cm} in ~\cref{tab:sota} is a LaViLa(L) model with an extra Transformer decoder, which is trained with only InfoNCE loss on video and sentence embeddings. Without object-awareness and hard sampling, it gets better accuracy on EGTEA and better nDCG on EK100 due to more parameters added, but falls behind on EgoMCQ and mAP on EK100. 
Applying hard negative sampling (Ours\ddag) and object-aware training (Ours) brings improvement across the board. The most obvious boost comes from inducing object-awareness, bringing 1.5\% improvement on average. And the results could be further improved by obtaining better pseudo-boxes, as the magnitude of boost from learning objects is closely related to the box quality (as shown in~\cref{tab:better_box}).

\paragraph{Episodic memory.}
Results on EgoNLQ and EgoMQ are shown in \cref{tab:nlq}. These two tasks test the video understanding on several-minutes long videos. In these experiments, trained video and text backbones are used as feature extractor, and extra modules are trained on the long feature sequences for natural language querying and memory querying. Therefore, the richer the information encoded in the features, the better the results will be.  We list the models trained on the same amount of video and text data in black, results show that our model are better than the previous SOTA on all the metrics in the two tasks. This is because features from the object-aware model have captured more object information in the clips, thus enabling better precision and recall on localization and retrieval.  We also include the InternVideo~\cite{wang2022internvideo} and NaQ~\cite{ramakrishnan2023naq} which are trained on more video or text data in the table for completeness.
\input{tables/egonlq}

\subsection{Evaluating Object Grounding}
\paragraph{Qualitative results on EgoClip.}
In \cref{fig:viz} we show the grounding results on EgoClip after training as compared to the supervision from the 100DOH detector. After training, the predictions from our model find the missing objects because we do not penalize extra bounding boxes predicted, but select the active object/object of interest through matching noun embeddings. It also learns temporal association of object bounding-boxes as the result of using the same object summary vector to predict the boxes over all the frames (as in \cref{eq:concat}); the summary vector attends to visually similar features corresponding to the same object across the clip without any explicit supervision for temporal consistency. 

\begin{figure}[h]
\centering
\includegraphics[width=\linewidth]{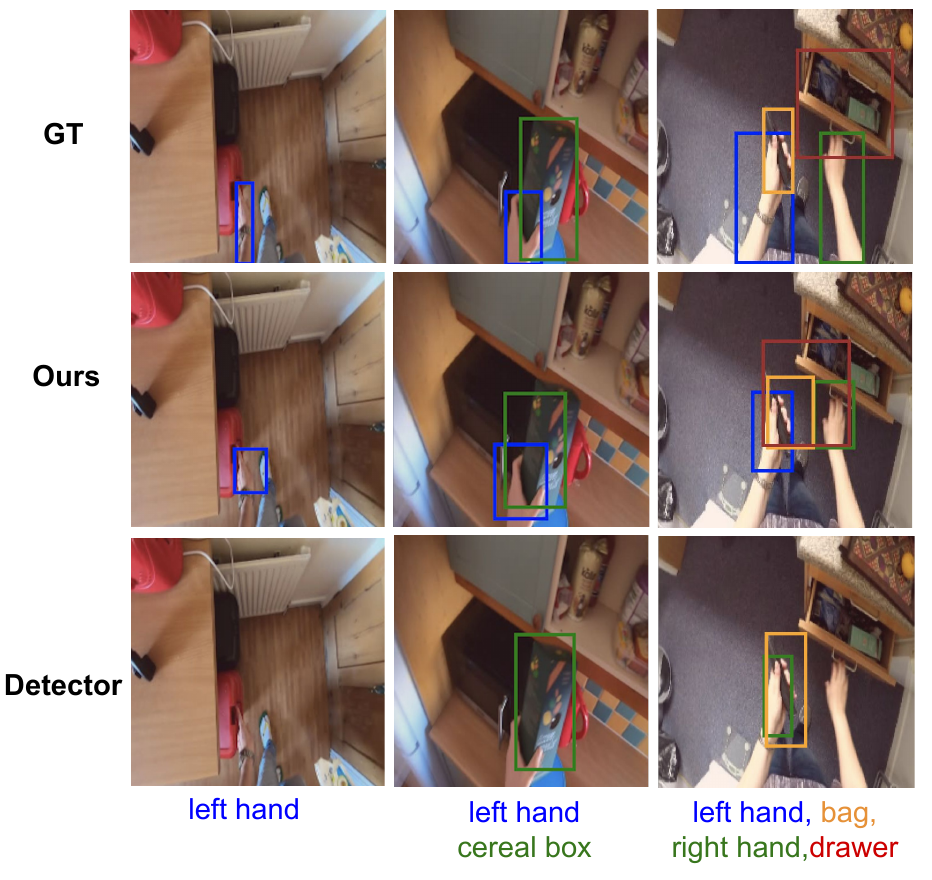}
\vspace{-5mm}
\caption{\textbf{Grounding visualization on the Epic-Kitchens-VISOR val split.} When operating at the same resolution, our model shows better grounding ability on hands and objects compared to the baseline 100DOH detector~\cite{100doh} (with GT matching) used for training. Note that the low IoU on \textcolor{blue}{hands} on the third column is a result of the GT `hand' segmentation mask covering the arm by definition, while the detector and our model are trained to localize only palm and fingers.}
\label{fig:visor}
\end{figure}

\vspace*{-2mm}
\paragraph{Quantitative and  qualitative results on VISOR.}
In \cref{tab:visor} we show the text-grounding results on VISOR. We take the predictions from the hand-object detector~\cite{100doh} as our baseline. Since the predictions only detect hands and objects without an object class name, we associate the predicted boxes with ground-truth boxes in VISOR in two ways: (a)~\textbf{random}: we assign predicted object boxes to GT object boxes randomly, (b)~\textbf{GT matching}: we use Hungarian matching to find the predicted boxes with highest IoU against the GT object boxes. However, even when matched using GT information, the baseline detector does not achieve a high accuracy due to poor recall. Results show grounding ability of our model is 40\% better than the baseline using only weak supervision from the video descriptions. We also show the qualitative results on detecting different number of hands and objects in \cref{fig:visor}, our model has a much higher recall when compared to baseline detector when operating at the same resolution.

\input{tables/visor}

%% file: tables/loss.tex
\begin{table}[ht]
\centering
\renewcommand*{\arraystretch}{1.2}
\resizebox{\linewidth}{!}{%
\begin{tabular}{ccccccc}
\hline
\multirow{2}{*}{\textbf{Losses}}   & \multicolumn{2}{c}{\textbf{EgoMCQ}} & \multicolumn{2}{c}{\textbf{EK100-MIR}}                                                                   & \multicolumn{2}{c}{\textbf{EGTEA}} \\
                                   & Inter             & Intra            & \begin{tabular}[c]{@{}c@{}}Avg \\ mAP\end{tabular} & \begin{tabular}[c]{@{}c@{}}Avg \\ nDCG\end{tabular} & Top1             & Mean            \\ \hline
$L_{Ego}$                          & 93.7             & 61.8            & 35.9                                               & 36.6                                             &  44.9                &   37.6              \\
$L_{Ego}$ + $L_{box}$              &  94.2             & 62.7            &  36.7                                              &  37.4                                           & 45.3                 &  38.5          \\
$L_{Ego}$ + $L_{word}$   &   93.7           &   61.9         & 36.7                &   37.4       &   45.8              &   38.1              \\
$L_{Ego}$ + $L_{box}$ +$L_{word}$  & \textbf{94.5}           &\textbf{63.0 }           & \textbf{37.5}                                             &  \textbf{37.8}                                             &\textbf{46.6}             & \textbf{39.1  }            \\ \hline
\end{tabular}%
}
\vspace{1mm}
\caption{\textbf{Ablation on training objectives for zero-shot transfer tasks.}  Introducing the object-awareness by having box and word supervision helps the model to achieve better transfer results on EK100-MIR and EGTEA.}
\vspace{-2mm}

\label{tab:loss}
\end{table}

%% file: tables/box_ablate.tex
\begin{table}[]
\centering
\renewcommand*{\arraystretch}{1.3}
\resizebox{0.48\textwidth}{!}{%
\begin{tabular}{c|cc|cc|cc|c}
\hline
 \multirow{3}{*}{\textbf{\begin{tabular}[c]{@{}c@{}}Detector \\ input res\end{tabular}}} & \multicolumn{2}{c|}{\textbf{EgoMCQ}}            & \multicolumn{2}{c|}{\textbf{EK}}                                                                                                           & \multicolumn{2}{c|}{\textbf{EGTEA}}    &    \textbf{VISOR}   \\
                                 &                                                                                      \multirow{2}{*}{Inter} & \multirow{2}{*}{Intra} & \multirow{2}{*}{\begin{tabular}[c]{@{}c@{}}Avg \\ mAP\end{tabular}} & \multirow{2}{*}{\begin{tabular}[c]{@{}c@{}}Avg\\  nDCG\end{tabular}} & \multirow{2}{*}{Top1} & \multirow{2}{*}{Mean} &  \multirow{2}{*}{Loc Acc}\\
                                 &                                                                                         &                        &                        &                                                                     &                                                                      &                       &                       \\ \hline
 256p                     &94.2   & \textbf{63.2}  & 35.7  & 34.6      & 42.0  & 36.0       & -           \\ \hline
 256p                   & \textbf{94.5}  & 63.0   & 36.9  & 37.0      & 44.3   & 38.9      & 68.2           \\
 640p                   & \textbf{94.5}   & 63.0   & \textbf{37.5}      & \textbf{37.8}  & \textbf{39.1}         & \textbf{46.6}    & \textbf{78.7}        \\ \hline
\end{tabular}%
}
\vspace{1mm}
\caption{\textbf{Ablation on box quality for zero-shot transfer tasks.} We extract boxes using 256p and 640p images as input to the detector respectively, resulting in boxes of different qualities as supervision in training.}
\vspace{-3mm}
\label{tab:better_box}
\end{table}

%% file: tables/n_query.tex
\begin{table}[h!]
\centering
\vspace{2mm}
\renewcommand*{\arraystretch}{1.2}
\resizebox{0.95\linewidth}{!}{%
\begin{tabular}{ccccccc}
\hline
\multirow{2}{*}{\textbf{\# Obj Queries}} & \multicolumn{2}{c}{\textbf{EgoMCQ}} & \multicolumn{2}{c}{\textbf{EK100-MIR}}                                                                   & \multicolumn{2}{c}{\textbf{EGTEA}} \\
                                            & Inter             & Intra            & \begin{tabular}[c]{@{}c@{}}Avg \\ mAP\end{tabular} & \begin{tabular}[c]{@{}c@{}}Avg \\ nDCG\end{tabular} &  \multicolumn{1}{c}{Top1} & \multicolumn{1}{c}{Mean}                        \\ \hline
4                                           & 94.1              & 62.8           & 37.8                                                &   37.6                                             &  45.9     &    38.1                   \\
8                                           & \textbf{94.5}              & 62.7           &  37.7                                                &\textbf{38.0}                                             &  45.5     &    37.9                   \\
12  & \textbf{94.5}           &\textbf{63.0}           &\textbf{37.5}                                             &  \textbf{37.8}                                             &\textbf{46.6}             & \textbf{39.1} \\ \hline
\end{tabular}%
}
\vspace{2mm}
\caption{\textbf{Ablation on the number of object queries  for zero-shot transfer tasks.} The number of queries do not have a big impact on EK100. Larger number of queries shows a boost on mean-class accuracy on EGTEA, and smaller number of queries is better on intra-video accuracy.}
 \vspace{-2mm}

\label{tab:nquery}
\end{table}

%% file: tables/sota.tex
\begin{table*}[]
\renewcommand*{\arraystretch}{1.1}

\resizebox{\textwidth}{!}{%
\begin{tabular}{ccccc|cc|cccccc|cc}
\hline
\multirow{3}{*}{\textbf{Method}} & \multirow{3}{*}{\textbf{Backbone}} & \multirow{3}{*}{\textbf{\begin{tabular}[c]{@{}c@{}}Batch\\ size\end{tabular}}} & \multirow{3}{*}{\textbf{\begin{tabular}[c]{@{}c@{}}Object\\ aware\end{tabular}}} &   \multirow{3}{*}{\textbf{\begin{tabular}[c]{@{}c@{}}Hard\\neg\end{tabular}}} & \multicolumn{2}{c|}{\textbf{EgoMCQ}}                        & \multicolumn{6}{c|}{\textbf{EK100-MIR}}             & \multicolumn{2}{c}{\textbf{EGTEA}}                    \\
                                 &                                    &                                         &                  &                        & \multirow{2}{*}{Inter-video} & \multirow{2}{*}{Intra-video} & \multicolumn{3}{c}{mAP} & \multicolumn{3}{c|}{nDCG} & \multirow{2}{*}{Top1-Acc} & \multirow{2}{*}{Mean-Acc} \\
                                 &                                    &                                                                                  &                              &                &     &         & V-T    & T-V    & Avg   & V-T     & T-V    & Avg    &                           &                           \\ \hline
EgoVLP    & TSF-B  & 512    & \xmark   & \cmark         & 90.6           & 57.2    & 26.0   & 20.6   & 23.3  & 28.8    & 27.0   & 27.9   & 17.6                         & -                         \\
LaViLa    & TSF-B   & 1024  & \xmark  & \xmark          & 93.8           & 59.9    & 35.1   & 26.6   & 30.9  & 33.7    & 30.4   & 32.0   &    -   & 28.9                         \\
LaViLa    & TSF-L   & 1024  & \xmark  & \xmark  & \textbf{94.5}          & 63.1    & 40.0   & 32.2   & 36.1  & 36.1    & 33.2   & 34.6   & 40.1   & 34.1                      \\
LaViLa*   & TSF-L   &  1024  & \xmark  & \xmark         & 94.2   & \textbf{63.2}   & 39.7   & 31.7   & 35.7  & 36.1    & 33.2   & 34.6   & 42.0   & 36.0                      \\\hline
Ours*\dag  & TSF-L   & 128   & \xmark  & \xmark         & 93.7           & 60.5    & 39.7   & 30.3   & 35.0  &  37.3   & 34.5   & 35.9   & 44.8   & 36.3                      \\
Ours*\ddag & TSF-L   & 128   & \xmark  & \cmark         & 93.7           & 61.8    & 40.7   & 31.1   & 35.9  &  38.3   & 35.0   & 36.6   & 44.9   & 37.6                      \\
Ours*      & TSF-L   & 128   & \cmark  & \cmark   & \textbf{94.5}        & 63.0    & \textbf{42.3}   &\textbf{32.7}   & \textbf{37.5}  &\textbf{39.3}   & \textbf{36.2}   &\textbf{37.8}   & \textbf{46.6}   & \textbf{39.1}                     
\\\hline
\end{tabular}%
}
\vspace{2mm}
\caption{\textbf{Comparison to SOTA results on zero-shot transfer to EgoMCQ, EK100-MIR and EGTEA.} We compared to EgoVLP and LaVILA, two previous SOTA models pre-trained on EgoClip. Our object-aware model has achieved comparable results on multiple-choice questions on EgoMCQ, and SOTA results on multi-instance retrieval task on EpicKicthens and action classification on EGTEA. \scriptsize{Model without * use center cropping in evaluation, while  * denotes the usage of resizing instead of cropping.  Ours\dag\hspace{0.5mm} and Ours\ddag\hspace{0.5mm}  stands for different variants of our model depending on whether object-aware losses and hard negative sampling is used in training.}}
\label{tab:sota}
\vspace{-3mm}
\end{table*}

%% file: tables/hand_ablate.tex
\begin{table}[]
\centering
\resizebox{0.45\textwidth}{!}{%
\begin{tabular}{cc|cc|cc|cc}
\hline
\multirow{3}{*}{\textbf{Method}} & \multirow{3}{*}{\textbf{\begin{tabular}[c]{@{}c@{}}Predicted\\ Boxes\end{tabular}}} & \multicolumn{2}{c|}{\textbf{EgoMCQ}}            & \multicolumn{2}{c|}{\textbf{EK}}                                                                                                           & \multicolumn{2}{c}{\textbf{EGTEA}}            \\
                                 &                                                                                     & \multirow{2}{*}{Inter} & \multirow{2}{*}{Intra} & \multirow{2}{*}{\begin{tabular}[c]{@{}c@{}}Avg \\ mAP\end{tabular}} & \multirow{2}{*}{\begin{tabular}[c]{@{}c@{}}Avg\\  nDCG\end{tabular}} & \multirow{2}{*}{Mean} & \multirow{2}{*}{Top1} \\
                                 &                                                                                     &                        &                        &                                                                     &                                                                      &                       &                       \\ \hline
\multirow{2}{*}{\textbf{Ours}}   & obj                                                                                 & 94.0             &   62.1      &  36.8                                                                   &    37.1                                                                  &   45.1                    &     37.6                  \\
                                 & hand+obj                                                                            & \textbf{94.5}              & \textbf{63.0}              & \textbf{37.5}                                                           & \textbf{37.8}                                                            & \textbf{46.6}             & \textbf{39.1}             \\ \hline
\end{tabular}%
}
\vspace{2mm}
\caption{\textbf{The impact of having hand boxes in the box prediction.} Having hand boxes in the prediction helps on all the zero-shot evaluation benchmarks.}
\label{tab:hand_box}
\vspace{-3mm}

\end{table}

%% file: tables/egonlq.tex


\begin{table}[htp]
\centering
\renewcommand*{\arraystretch}{1.5}

\resizebox{1.0\linewidth}{!}{%
\begin{tabular}{ccccccccc}

\hline
\textbf{Method}    & \begin{tabular}[c]{@{}c@{}}\textbf{Batch} \\ \textbf{size}\end{tabular} & \multicolumn{4}{c}{\textbf{EgoNLQ}}                                  & \multicolumn{3}{c}{\textbf{EgoMQ}}                                                \\
          &                                                       & \multicolumn{2}{c}{mIOU@0.3} & \multicolumn{2}{c}{mIOU@0.5} & \multirow{2}{*}{R1@0.5} & \multirow{2}{*}{R5@0.5} & \multirow{2}{*}{mAP} \\
          &            & R1            & R5           & R1           & R5            &                         &                         &                      \\ \hline
SlowFast  & -          & 5.5          & 10.7        & 3.1         & 6.6          & 25.2                    & 46.2                    & 6.0                  \\
EgoVLP    & 512        & 10.8         & 18.8        & 6.8         & 13.5         & 30.1                    & 52.0                    & 11.4                 \\
LaViLa(B) & 1024       & 10.5         & 19.1        & 6.7         & 13.6         & -                       & -                       & -                    \\
LaViLa(L) & 1024       & 12.1         & 22.4        & 7.3         & 15.4         & 32.5                    & 56.1                    & 13.4                 \\ \hline
Ours      & 128        &\textbf{13.2}& \textbf{23.3} &\textbf{7.9}& \textbf{15.6}& \textbf{33.4}           & \textbf{56.7}           & \textbf{16.0}      \\ \hline

\color{gray}VideoIntern~\cite{wang2022internvideo}     & \color{gray}14k      &\color{gray}16.5  & \color{gray}23.0  &\color{gray} 10.1  & \color{gray}16.1 & -          & -          & \color{gray} 23.6      \\ 
\color{gray}ReLER + NaQ~\cite{ramakrishnan2023naq}      & \color{gray}2048        & \color{gray}19.3 & \color{gray}23.6  & \color{gray}11.6   &\color{gray}15.5          & \color{gray}-          & -    & -    \\ \hline
\end{tabular}
}
\vspace{1mm}
\caption{\textbf{Comparison to SOTA results of fine-tuning on Ego-NLQ and EgoMQ.} Our object-aware model encodes richer information in the visual representations, hence obtaining better results on all the metrics in NLQ and MQ task in Ego4D episodic memory benchmark. \color{gray}We list other SOTA models (in grey) trained with more video and text data for completeness.}
\label{tab:nlq}
\vspace{-3mm}

\end{table}

%% file: tables/visor.tex
\begin{table}[]
\centering
\renewcommand*{\arraystretch}{1.3}

\resizebox{\linewidth}{!}{%
\begin{tabular}{cccc}
\hline
\textbf{Model}                        & \multicolumn{1}{l}{\textbf{Object Assignment}} & \textbf{\# queries} & \textbf{Loc Accuracy} \\ \hline
\multirow{2}{*}{Detector~\cite{100doh}} & Random                                & -                    &    37.1                   \\
                                      & GT matching                                   & -                    &  41.3                     \\ \hline
\multirow{3}{*}{Ours}                 & \multirow{3}{*}{Predicted}                    & 4                    &\textbf{82.3}                  \\
                                      &                                               & 8                    & 81.2                  \\
                                      &                                               & 12                   & 78.7                  \\ \hline
\end{tabular}%
}
\vspace{1mm}
\caption{\textbf{In-contact object localization accuracy on VISOR.} Our model does better on in-contact object localization after weakly-supervised training compared to the baseline (the Detector with Random or GT object assignment), which is the source of supervision in our pre-training. The improvement is due to increased recall of our model over the baseline.}
\label{tab:visor}
\vspace*{-3mm}
\end{table}

%% file: sections/conclusion.tex
\section{Conclusion}
In this paper, we introduce a method to learn object-aware ego-centric video representations using noisy supervision from pre-trained hand-object detectors.
The object-representations so learned show strong zero-shot transfer across various downstream tasks and datasets, mirroring the performance improvement from  object-aware training on   images~\cite{zeng2021xvlm}.
The model uses standard neural modules (i.e., transformers), and does not require any object boxes or detectors as input during inference, making it widely applicable as a drop-in replacement for training video-language models.
Even though the model is trained with sparse and noisy object supervision at the frame-level (without temporal associations), during inference dense temporal bounding-box tracks and category predictions can be obtained, which are superior to the predictions from the base hand-object detector used for training. 
There are several avenues for improvement.
Our model uses the pre-trained video encoder operates at a small resolution $224{\times}224$ which makes detecting small objects difficult. Further, four frames are sampled uniformly from the clip regardless of its length which can cause difficulties due to temporal aliasing.
Nevertheless, we hope our work will inspire further research in learning transferable object-aware representations for videos.

\paragraph{Acknowledgements.} This research is funded by a Google-DeepMind Graduate
Scholarship, a Royal Society Research Professorship, and EPSRC Programme
Grant VisualAI EP/T028572/1.

%% file: supp/box_region.tex
\section{Discussion: Text-Region Alignment}
One of the most related work to our paper is~\cite{yao2021filip}, Yao et al.\ train dual encoders to align image patches and textual words. Fine-grained pre-training helps the model to achieve better results on image classification and image-text retrieval.
GLoRIA~\cite{huang2021gloria} is another similar work on medical image recognition, where they show region-word matching is a more label-efficient pre-training method compared to image-sentence matching on retrieval, classification and segmentation. While our work focuses on egocentric videos and utilizes detections from hand-object detector to supervise the training for alignment. This is because scenes in egocentric videos are often crowded and objects are prone to be heavily occluded. 
Boxes from off-the-shelf detectors are easy to obtain and can largely ease the training process. Furthermore, as an important factor in first-person videos, hands are not often mentioned in the narrations; explicit supervision helps the model to focus on the motion during training.
Similarily, another line of work~\cite{zeng2021xvlm,dou2022fiber} pre-trains a vision-language model to predict object boxes, but relies on manually labeled ground-truth. While our model can be trained with imperfect supervision.  
Other works~\cite{rao2022denseclip,zhong2022regionclip} train models to do pixel-text or region-text alignment for open-vocabulary detection or segmentation.

%% file: supp/Implementation.tex
\section{Implementations}

\subsection{Training}
Given a video clip, we uniformly sample 4 frames from the clip, and resize the image to $224\times224$ without cropping, color jittering is applied as data augmentation. We use the 3.8M video clips from EgoClip for training. Each clip is paired with its original narration from EgoClip and the rephrased ones from LaViLA~\cite{zhao2022lavila}. The additional pseudo-labelled video clips from LaViLA are not used.

\subsection{Evaluation}

\paragraph{EgoMCQ.}
EgoMCQ dataset is a multiple choice question dataset built on Ego4D. 
Given one narration as question, the model is tasked to find the paired video clip from 5 candidates. 
It has 39k questions in total, which are categorized into `inter-video' and `intra-video' multiple-choice questions. 
There are 24k questions in the “inter-video” split, where the candidates are from different videos. 
The “intra-video” split has 15K questions, where the candidates are from the same video. The average temporal gap between the intra-video candidates is 34.2 seconds. We sample 4 frames uniformly from each clip and resize them to $224\times224$ as input in evaluation. 

\paragraph{EpicKitchens-MIR.}
Epic-Kitchens Multiple Instance Retrieval is a dataset from Epic-Kitchens 100 for video-text and text-video retrieval.
Given a query video/caption, 
the task is to rank the instances from the other modality such that higher-ranked instances are more semantically relevant to the query. 
We use the val split for zero-shot transfer evaluation, which contains 9668 video-caption pairs. 
The captions are in the format of 'verb + noun', with totally 78 verb classes and 211 noun classes. In evaluation, We sample 16 frames uniformly across the clip, and resize frames to $224\times224$ as input. Mean Average Precision (mAP) and normalized Discounted Cumulative Gain (nDCG) are used as evaluation metrics. 

\paragraph{EGTEA.}
EGTEA contains 28 hours of cooking activities from 86 unique sessions of 32 subjects. We evaluate the model on action classification and use top-1 accuracy and mean-class accuracy as metrics. 
The descriptions of 106 action classes are encoded into text embeddings using the text encoder. 
We compute the similarity score between every video embedding and the 106 text embeddings, and take the text embedding with the highest similarity score as the predicted class. 
Evaluation is done on its first test split with 2022 instances. We uniformly sample 10 clips from the full span of one video instance, each has 16 frames with a temporal stride of 2. We resize the frames to $224\times224$ as input to the model. 
For each video instance, 
we predict logits for 10 clips and then max-pool the logit as the final prediction. 

\paragraph{EgoNLQ.}
Given a video clip and a query expressed in natural language, 
the task is to localize the temporal window within all the video history where the answer to the question is evident. 
We evaluate the model on the val split covering 45-hour videos, with 0.3k clips and 3.9k queries. We follow ~\cite{egovlp} and extract all the video and text embeddings using our model, and input them to VSLNet~\cite{zhang2021vslnet} for fine-tuning on EgoNLQ. 
The evaluation metrics are based on the overlap of top-1 or top-5 predicted temporal windows with the ground-truth at IoU thresholds of 0.3 and 0.5. 

\paragraph{EgoMQ.}
In this task is a natural language grounding task, where activities are used as queries to find responses consisting of all temporal windows where the activity occurs in a video. There are 13.6 training instances from 1.5k clips and 4.3k validation instance from 0.5k clips. We extract all the video features using our model as input, and train a VSGN~\cite{zhao2021vsgn} to perform the task. We report mAP and recall as evaluation metrics.

\paragraph{VISOR.}
VISOR is a dataset built on videos from Epic-Kitchens 100 for segmenting hands and active objects in egocentric videos. We re-propose VISOR for a in-contact hand-object grounding by 1) converting the segmentation masks to bounding boxes 
2) filtering out not-in-contact objects in the frames. 
Given a list of names (hand + in-contact objects), the model is tasked to predict a bounding box for each instance. 
We do evaluation on the val split with 7,747 images, 182 entity classes from 4 videos. After filtering, the model has 1.4 hands and 0.9 objects per frame on average. We resize each image to $224\times224$ and repeat it for 4 times along the temporal dimension to make it a 4-frame clip as input to the model. 
For hands, we always use the first hand query for left hand box prediction, and the second hand query for right hand box predication, as we find hand queries have learned to specify without explicit supervision. For objects, we match the text embedding of object names with the predicted object embeddings for grounding as in training. 

For the baseline image detector, we also resize the shorter side of the image to 224p as input for fair comparison. The detector produces two types of output: hand boxes with 'left' and 'right' labels, and object boxes without object class. Since there is no specific grounding predicted by the detector, we conduct two types of matching in our evaluation:
\begin{itemize}
\item \textbf{Random matching:} The predicted object boxes are randomly assigned to ground-truth objects
\item \textbf{Hungarian matching:} We compute the IoU between predicted boxes and ground-truth boxes, and apply Hungarian matching for grounding. 
\end{itemize}


%% file: supp/datasets.tex
\section{Statistics}
\subsection{Grounded Nouns in EgoClip}

The Ego4D taxonomy dictionary~\cite{grauman2022ego4d} is a thesaurus that records meaningful nouns/verbs in Ego4D narrations, it has 581 noun groups with 1610 nouns. We match all the single words and two-word phrases in the narrations with nouns in the dictionary to extract the nouns from the narrations. We remove nouns that refer to the background or someone who is holding the camera, including: `man', `woman', `person', `lady', `they', `ground', `camera', `table', and `leg'. We also remove nouns related to 'hand' because we use hand supervision from the object-hand detector instead of the narrations.  As a result, we find 5,020,303 nouns from 3,847,723 narrations in training. Below, we plot a histogram of the top 45 nouns in EgoClip. 
\begin{figure}
    \centering
\includegraphics[width=\linewidth]{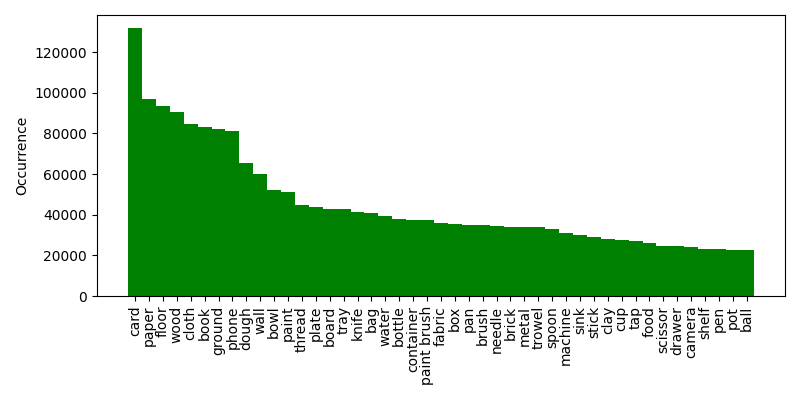}
    \caption{\textbf{The distribution of op 45 nouns in EgoClip.}}
    \label{fig:my_label}
\end{figure}

\subsection{Out-of-Distribution Nouns in VISOR}

We compare the 1610 nouns in the pre-training dataset EgoClip, and 411 nouns in the downstream grounding dataset VISOR. There are 250 noun words/noun phrases in VISOR that have not appeared in the pre-training. Some are new combinations with an additional adjective, e.g., small bread, hot water, aluminium foil. Some are objects that have not appeared in the pre-training, e.g., basil, scale, drainer.  As results shown in ~\cref{tab:seen_unseen}, the localization accuracy is 48.4\% on unseen concepts and 70.9\% on seen concepts.  The reason that our model is able to ground some of the unseen concepts is probably: 1) Some unseen nouns/phrases have similar semantic meaning with the seen ones, hence the word embeddings can be similar. e.g., hot water and water. 2) When there is no other distractive object in the scene, all the object queries localize the same object that is in contact with the hand. In this case, the proposed box can always be matched to the object of interest no matter whether it is seen or not.

\begin{table}[]
\centering
\renewcommand*{\arraystretch}{1.2}

\resizebox{0.8\linewidth}{!}{%
\begin{tabular}{cccc}
\hline
        & \textbf{Unseen} & \textbf{Seen} & \textbf{Overall} \\ \hline
Occurrence & 2,041         & 15,800   & 17,841      \\ \hline
Localization Acc & 52.8\%           & 82.0\%   & 78.7\%       \\ \hline
\end{tabular}%
}
\vspace{1mm}
\caption{\textbf{Localization accuracy on seen and unseen nouns/phrases on VISOR.}}
\label{tab:seen_unseen}
\end{table}